\documentclass[a4paper]{article}
\usepackage{soul}
\usepackage{INTERSPEECH2021}
\usepackage{cite}
\usepackage{caption}
\usepackage{subcaption}
\usepackage{multirow}
\usepackage{float}
\usepackage{bbding}
\usepackage{enumitem}
\usepackage{hyperref}
\usepackage{xcolor,graphicx}
\usepackage{balance,subcaption}
\usepackage{nicefrac}

\title{Partially-Connected Differentiable Architecture Search\\for Deepfake and Spoofing Detection}
\name{Wanying Ge, Michele Panariello, Jose Patino, Massimiliano Todisco and Nicholas Evans}

\address{
  EURECOM, Sophia Antipolis, France}
\email{firstname.lastname@eurecom.fr}

\begin{document}

\maketitle
\begin{abstract}
This paper reports the first successful application of a differentiable architecture search (DARTS) approach to the deepfake and spoofing detection problems.  An example of neural architecture search, DARTS operates upon a continuous, differentiable search space which enables both the architecture and parameters to be optimised via gradient descent. Solutions based on partially-connected DARTS use random channel masking in the search space to reduce GPU time and automatically learn and optimise complex neural architectures composed of convolutional operations and residual blocks. Despite being learned quickly with little human effort, the resulting networks are competitive with the best performing systems reported in the literature. Some are also far less complex, containing 85\% fewer parameters than a Res2Net competitor.

\end{abstract}
\vspace{0.1cm}

\section{Introduction}

Compared to automatic speaker verification for which the research history is decades long, research in deepfake or spoofing detection is relatively embryonic. While recent years have seen rapid progress, front-end feature extraction as well as back-end classification approaches are still evolving~\cite{SahidullahDTKEYL19}. Early work is characterised by a focus on front-end feature engineering, namely the design of parameters or representations which capture the tell-tale signs of manipulated or synthesized speech signals and which help to distinguish these from bona fide speech~\cite{todisco2016cqcc, zhang2020oneclass}.  More recently, greater attention has been paid to the back-end classifier design.  Like all fields of speech processing, deep neural network architectures are the classifier of choice~\cite{li2019butterfly, Lai2019assert}.
    
The use of end-to-end (E2E) processing, whereby hand-crafted and manually optimised components are replaced with automatically designed and optimised substitutes, has attracted growing attention.  Thus far, E2E developments extend mostly to the front-end components~\cite{jung2019rawnet,ravanelli2018sinc}.  While back-end components can be similarly optimised, this usually extends only to the network \emph{parameters}; the network \emph{architecture} itself is almost always still hand-crafted. Inspired by original work in~\cite{stanley2002NEAT, daniel2017evolving}, our first attempt to harness the potential of fully E2E processing~\cite{valenti2018end} explored the use of neuro-evolution for augmenting topologies (NEAT). While NEAT is successful in learning network architectures automatically, performance was found to be far from the state of the art, while computational complexity was found to be prohibitive. 
Whereas more efficient NEAT implementations are reported in the literature~\cite{stanley2009hyper-neat, risi2011enhancingNEAT}, we have instead turned to powerful and efficient alternatives with proven potential in speech-related tasks.

We have explored the use of neural
architecture search (NAS), originally proposed in~\cite{zoph2016neural}.  NAS solutions are based upon an architecture \emph{search space}, a \emph{search strategy} and an \emph{evaluation strategy}~\cite{elsken2019nassurvey}.  A search space contains a set of candidate \emph{operations}. 
Using some performance criteria, an architecture is selected from this space and further optimised. The particular variant of NAS known as differentiable architecture search (DARTS)~\cite{liu2018darts}, enables the selection of candidate operations, and hence the architecture, from a search space with continuous and learnable weights. DARTS models can be optimised with backpropagation in the usual manner with hardware acceleration. The network is designed automatically by optimising the operations contained within architecture building blocks referred to as \emph{cells}.  Candidate operations, including convolutional operations, pooling layers, and residual connections among others, are selected during an initial search phase, before the resulting cells are stacked together to build a deeper architecture which is then further optimised. 
The resulting networks resemble the current state of the art in anti-spoofing, hence our adoption of DARTS in this work.

This paper reports our use of a particular variant of DARTS known as partial channel connections (PC-DARTS)~\cite{xu2019pc} for anti-spoofing.  We show how partial channel connections, which deliver substantial savings in both computational complexity and memory, enable the automatic learning of a neural network based solution to anti-spoofing. Both the network architecture and parameters are learned automatically with only minimal human input.  To the best of our knowledge, our work is both the first reported application of DARTS to anti-spoofing and the first reported application of PC-DARTS in \emph{any} field of speech processing. The remainder of the paper is organised as follows. Section 2 introduces the related work and objectives. The proposed system is reported in Section 3. Experiments and results are reported in Sections 4 and~5. Our findings and conclusions are reported in Section 6.

\section{Related work and objectives}

DARTS has already been applied successfully to speech and language tasks~\cite{Mo2020NAS-KWS, Chen2020darts-asr, Ding2020}. Its use for architecture search in a keyword spotting task is reported in~\cite{Mo2020NAS-KWS}.  Competitive results were obtained with a search space containing the regular operations used in ResNet.
A successful application to automatic speech recognition reported in~\cite{Chen2020darts-asr} showed promising results even when architecture search and training stages are performed using different language datasets.
The first application of DARTS to speaker verification is reported in~\cite{Ding2020} which shows that smaller, automatically learned solutions compare favourably to hand-crafted architectures.  
While results comparable to the state of the art are reported in both~\cite{Mo2020NAS-KWS} and~\cite{Ding2020}, both also report the necessary use of small batch sizes so that architecture search can be performed upon a single GPU.  

The first objective of our work is hence to determine whether neural architectures learned automatically with PC-DARTS can compete with hand-crafted networks. Second, we seek to determine the longer term scope for such networks to even outperform the current state of the art.  Third, we are interested to learn whether automatically learned and optimised solutions are more efficient. While not an objective of the current work, our hypothesis is also that PC-DARTS may yield less complex networks whose behaviour may be more easily \emph{explained}.

\section{PC-DARTS}

\begin{figure}[t]
         \centering
         \includegraphics[trim=0 2.75cm 0 0.5cm, clip,width=\columnwidth]{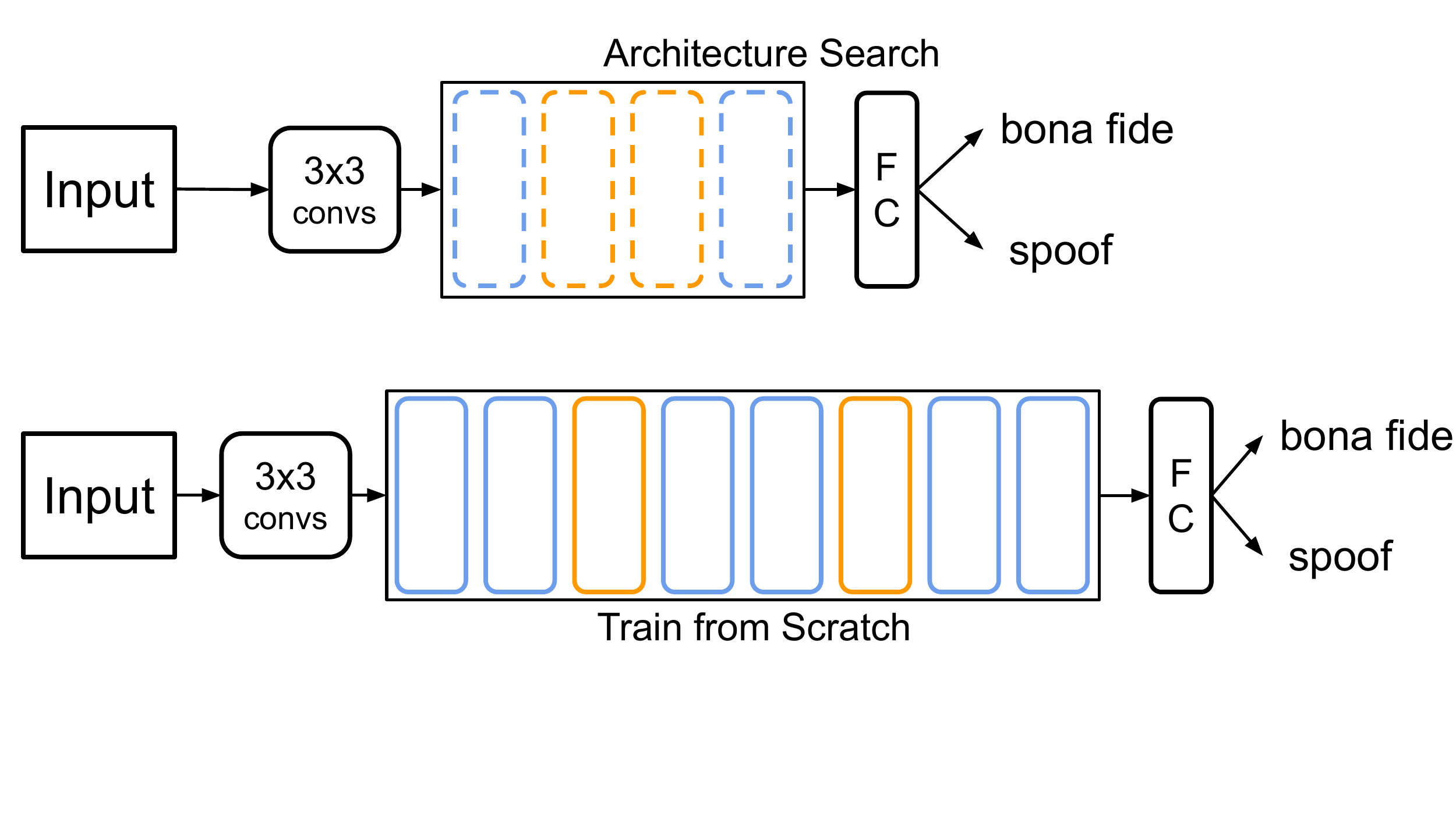}
         \caption{An illustration of architecture search and train from scratch. Architecture search optimises a stack of 2 normal cells (dashed blue) and reduction cells (dashed yellow). The train from scratch stage optimises a deeper network of stacked cells (solid blue and yellow). Only network parameters are optimised in the second stage; the cell architectures are those fixed during architecture search.}
         \label{fig:whole_system}
         \end{figure}

As illustrated in Figure~\ref{fig:whole_system}, DARTS encompasses a pair of learning stages referred to as \emph{architecture search} (top half) and \emph{train from scratch} (bottom half).
A key idea is to construct a complex network architecture from a pair of building blocks, referred to as \emph{cells} (blue and yellow blocks in Figure~\ref{fig:whole_system}), whose internal architecture and parameters are learned automatically. In contrast to other NAS approaches which search over a discrete set of candidate network operations, DARTS operates upon a relaxed, continuous search space. This makes the architecture representation itself \emph{differentiable}, meaning that it can be optimised in the usual manner via gradient descent and backpropagation with hardware acceleration. In the architecture search stage, the cell \emph{architecture parameters} are learned and fixed.  The train from scratch stage operates upon a deeper network formed from the stacking of a greater number of cells, thereby forming a deeper residual network.  The \emph{network parameters} are then re-optimised. The initial architecture search stage is computationally demanding.  The use of partial connections (PC-DARTS) provides a more efficient solution. Since neither DARTS, much less PC-DARTS are mainstream within the speech community, a brief overview of both is provided in the following.

\subsection{Searching for the Optimal Architecture} \label{section:arch_search}

         \begin{figure}[h]
         \centering
         \includegraphics[width=\columnwidth]{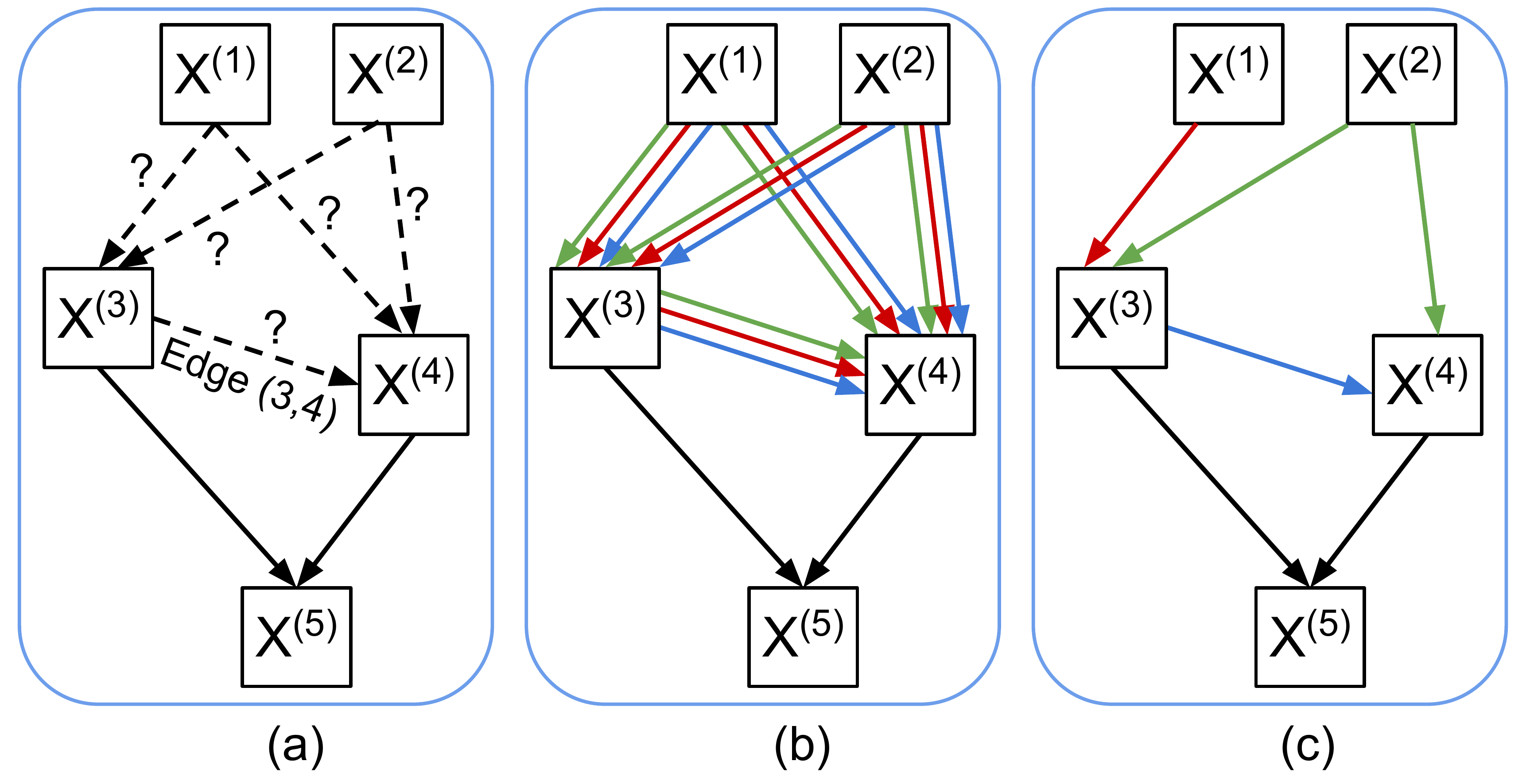}
         \caption{An illustration of architecture search: (a)~a neural cell with \(N=5\) nodes; (b) an illustration of the candidate operations performed on each edge that are optimised during architecture search; (c)~resulting optimised cell with \(K=2\) inputs to each intermediate node.}
         \label{fig:cell}
         \end{figure}

DARTS networks are constructed from the concatenation of multiple \emph{cells}.  An example is illustrated in Figure~\ref{fig:cell}.  Their internal architectures are learned automatically and dictate the sequence of operations performed upon input data in generating their output. 

Each cell contains \(N\) nodes, where each node 
\(\mathbf{x}^{\left(i\right)}\)
represents a feature map in tensor form. The first pair of nodes, \(\mathbf{x}^{\left(1\right)}\) and \(\mathbf{x}^{\left(2\right)}\), are the cell inputs and are connected to the outputs of the previous two cells.
Nodes \(\mathbf{x}^{\left(3\right)}\) to \(\mathbf{x}^{\left(N-1\right)}\), referred to as intermediate nodes, are computed from previous nodes with
operation \(o\) selected from the search space \(\mathcal{O}\) according to:
\begin{equation} \label{eq:node}
\mathbf{x}^{\left(j\right)}=\sum_{i<j}o^{\left(i,j\right)}\left(\mathbf{x}^{\left(i\right)}\right)
\end{equation}    
where \(o^{\left(i,j\right)}\) 
is the operation performed on edge $(i,j)$ that connects \(\mathbf{x}^{\left(i\right)}\) and \(\mathbf{x}^{\left(j\right)}\). Node \(\mathbf{x}^{\left(N\right)}\) is the cell output: its feature map is constructed from the concatenation of the feature maps corresponding to the full set of intermediate nodes.

In the architecture search stage, a linear combination of operations, denoted as $\Bar{o}$, is performed on edge $(i,j)$ according to some weight $\alpha_o^{\left(i,j\right)}$. 
The weights form a continuous search space through a softmax function:
\vspace{-0.15cm}
    \begin{equation} \label{eq:mixed_op}
    \Bar{o}^{\left(i,j\right)}\left(\mathbf{x}^{\left(i\right)}\right) = \sum_{o \in \mathcal{O}} \frac{\exp \left(\alpha_o^{\left(i,j\right)}\right)}{\sum_{o' \in \mathcal{O}} \exp \left(\alpha_{o'}^{\left(i,j\right)}\right)} \, o\left(\mathbf{x}^{\left(i\right)}\right)
    \end{equation}

Architecture search is hence reduced to the learning of a set of continuous variables \(\boldsymbol\alpha=\{\alpha^{\left(i,j\right)}\}\), where $\alpha^{\left(i,j\right)}$ is a vector of dimension $|\mathcal{O}|$. Both the {\it architecture parameters} \(\boldsymbol\alpha\) and the {\it network parameters} \(\boldsymbol\omega\) (e.g.\ the convolutional filter weights) can be jointly optimised through backpropagation. 
The goal is to determine the \(\boldsymbol\alpha\) which minimises the validation loss \(L_{val}\), where the optimal \(\boldsymbol\omega\) is determined by minimising the training loss \(L_{train}(\boldsymbol\omega,\boldsymbol\alpha)\):
    \begin{equation}
    \begin{aligned}
    &\min_{\boldsymbol\alpha} L_{val}(\boldsymbol{\omega}^*,\boldsymbol\alpha) \\
    &\text{s.t.}\;\;\boldsymbol{\omega}^* = \underset{\boldsymbol\omega}{\operatorname{argmin}}\; L_{train}(\boldsymbol\omega,\boldsymbol\alpha)
    \end{aligned}
    \end{equation}
When the search stage is complete, \(\Bar{o}^{\left(i,j\right)}\) is replaced with the single operation with the highest $\alpha_{o}^{\left(i,j\right)}$. The final cell architecture is obtained by retaining the set of \(K\) edges entering each intermediate node which have the highest weights $\alpha_{o}^{\left(i,j\right)}$, where \(K\) is a hyperparameter.  The remainder are discarded.

The search space \(\mathcal{O}\) proposed in~\cite{Ding2020} comprises: a \(3\times3\) separable convolution; a \(5\times5\) separable convolution; a \(3\times3\) dilated convolution; a \(5\times5\) dilated convolution; a skip connection; a \(3\times3\) average pooling; a \(3\times3\) max pooling; none (no connection).  The set of operations are used in defining two types of neural cells, namely \emph{normal} cells and \emph{reduction} cells.  
As illustrated to the base of Figure~\ref{fig:whole_system}, cells are stacked together to form the full, deeper residual network, with reduction cells being placed at the \(\frac{1}{3}\) and \(\frac{2}{3}\) depth positions of the total network depth (number of stacked cells). Feature map dimensions are fixed for the input and output of each normal cell. Reduction cells act to reduce the feature map dimensions by 50\% while doubling the number of channels.

\subsection{Partial Channel Connections and Edge Normalisation}

DARTS remains computationally demanding, especially in the architecture search stage. To improve efficiency, we used partial channel connections and edge normalisation~\cite{xu2019pc}. Partially-connected DARTS (PC-DARTS) delivers substantial savings in computation and memory. For a given edge \(\left(i,j\right)\), partial channel connections are formed from the element-wise multiplication of \(\mathbf{x}^{\left(i\right)}\) by a masking operator \(\mathbf{S}^{\left(i,j\right)}\) of the same dimension. The masking operator either \emph{selects} (multiplication by 1) or \emph{masks} (multiplication by 0) each channel in \(\mathbf{x}^{\left(i\right)}\):
\vspace{-0.15cm}
\begin{multline}
\Bar{o}^{\left(i,j\right)}\left(\mathbf{x}^{\left(i\right)}\right) =
\sum_{o \in \mathcal{O}} \frac{\exp{\left(\alpha_o^{\left(i,j\right)}\right)}}{\sum_{o' \in \mathcal{O}} \exp{\left(\alpha_{o'}^{\left(i,j\right)}\right)}}\, o\left(\mathbf{S}^{\left(i,j\right)} \odot \mathbf{x}^{\left(i\right)}\right)\\ + \left(1 - \mathbf{S}^{\left(i,j\right)}\right) \odot \mathbf{x}^{\left(i\right)}
\end{multline}
where $\odot$ indicates element wise multiplication.  A hyperparameter \(K_C\) is set to conserve a random fraction \(1/K_C\) of the available channels. Partial connections hence reduce the computational load by a factor $K_C$ while acting to regularise the choice of weight-free candidate operations (e.g., max pooling) in \(\mathcal{O}\) for a given edge\cite{xu2019pc}. There is hence a trade off between performance (smaller \(K_C\)) and efficiency (larger \(K_C\)). As a result of random channel sampling, the linear combination of operations $\Bar{o}^{\left(i,j\right)}$ for each node can become unstable under training. This issue is addressed by introducing a set of \emph{edge normalisation} parameters $\boldsymbol\beta$ which smooth node inputs according to:
\begin{equation}
\mathbf{x}^{\left(j\right)}=\sum_{i<j} \frac{\exp \left(\beta^{\left(i,j\right)}\right)}{\sum_{i'<j} \exp \left(\beta^{\left(i',j\right)}\right)} \; \Bar{o}^{\left(i,j\right)}\left(\mathbf{x}^{\left(i\right)}\right)
\end{equation}
where \(\beta^{\left(i,j\right)}\) is a learnable smoothing factor. The set of architecture parameters optimised by minimizing \(L_{val}\) now comprises both \(\boldsymbol\alpha\) and \(\boldsymbol\beta\).

\section{Experiments}

In this section we describe the experimental setup, the choice of front-end and our specific PC-DARTS configuration.

\subsection{Database, protocols and metrics}

All work reported in this paper was performed using the ASVspoof 2019 Logical Access (LA) database~\cite{wang2020asvspoof} which comprises the usual train, development and evaluation partitions. In the architecture search stage, a random selection of half the number of utterances for each class in the training partition, including bona fide and spoofed (A01-A06), is used to learn network parameters. The other half is used to learn architectures, namely one normal cell and one reduction cell. The cell architectures which produce the highest classification accuracy are then used in the train from scratch stage.

After the train from scratch stage, the performance of the resulting model is assessed using the full evaluation partition.
Performance is reported in terms of the pooled minimum normalised tandem detection cost function (min-tDCF)~\cite{kinnunen2018tdcf}
in addition to the pooled equal error rate (EER).
    
\subsection{Front-end}

Initial experiments showed that the application of neural architecture search to raw audio waveforms places excessive demands upon GPU memory, implying lower batch sizes and greater training time~\cite{tak2020end}. We hence used linear frequency cepstral coefficients (LFCCs) 
of 60 dimensions
encompassing static, delta and delta-delta coefficients. Features are extracted using 64~ms Hamming windows with a 16~ms shift and a 1024-point FFT. In order to improve generalisation,
frequency masking is applied according to the procedure described in~\cite{chen2020generalization} with a maximum of $12$ masked frequency channels per mini-batch.

\subsection{PC-DARTS}

As is customary\cite{xu2019pc}, we applied three convolutional layers of stride 2 to the input features in order to reduce resolution. Architecture search is performed using 4 neural cells (2 normal cells and 2 reduction cells) with 16 initial channels. Each cell has \(N=7\) nodes, and each intermediate node retains \(K=2\) inputs after search.

Training for the architecture search stage is performed for 50 epochs with a batch size of 64 using an Adam optimiser to learn both architecture and network parameters. Both are optimised by minimising the weighted cross-entropy loss between spoofed and bona fide data with a ratio \(1:9\). According to~\cite{xu2019pc,yu2020searching}, architecture parameters are not updated in the first 10 epochs. For the learning of architecture parameters we used a learning rate of 6e-4 and a weight decay of 0.001. For network parameters, we used an initial learning rate of 0.01 which is annealed down to 0.001 according to a cosine schedule.
Partial channel connections use a value of \(K_C=2\). When the architecture search stage is complete, network parameters \(\boldsymbol\omega\) are forgotten.  Only the normal and reduction cell architectures are then retained.

During the train from scratch stage, models are trained for 100 epochs with a batch size of 128 and an initial learning rate of 0.001. The drop-path rate~\cite{xu2019pc} is set to 0.2. We experimented with different numbers of stacked layers~(\(L\)) and initial channels~(\(C\)). The models are optimized with the same loss function as in the architecture search stage. The final scores are taken from the output for the bona fide class. 

All experiments reported in this paper were performed on a single NVIDIA GeForce RTX 2080 Ti GPU. Using the implementation available online\footnote{\href{https://github.com/eurecom-asp/pc-darts-anti-spoofing}{https://github.com/eurecom-asp/pc-darts-anti-spoofing}}, all results are reproducible with the same random seed and GPU environment.

\begin{figure*}
  \begin{subfigure}[t]{\columnwidth}
  \includegraphics[width=\columnwidth]{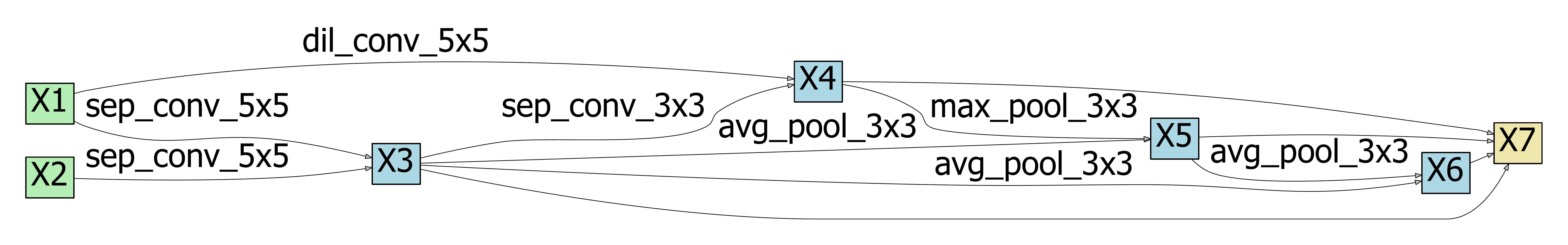}

    \caption{Normal cell}
    \label{fig:normal_cell}
  \end{subfigure}
  \hfill 
  \begin{subfigure}[t]{\columnwidth}
  \includegraphics[width=\columnwidth]{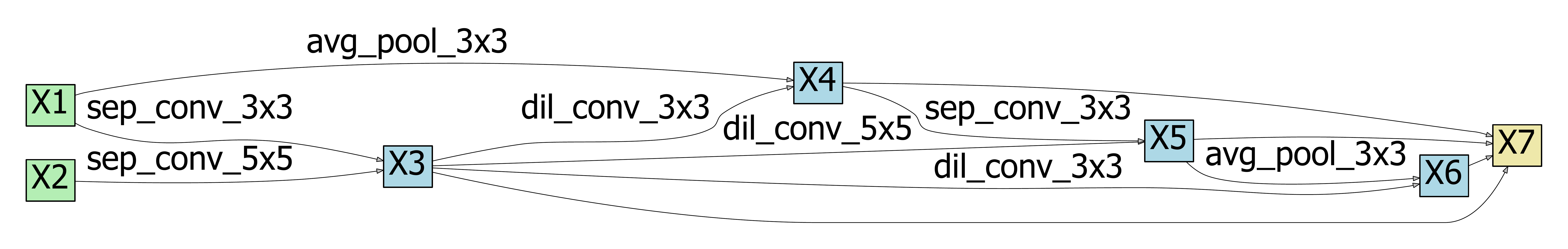}
    \caption{Reduction cell}
    \label{fig:reduction_cell}
  \end{subfigure}
  \centering
  \caption{An illustration of the (a) normal and (b) reduction cells resulting from architecture search. As illustrated in Figure~\ref{fig:whole_system}, they form the basic building blocks used to construct the architecture used in the train from scratch stage.}
  \label{fig:cellIllustrations}
\end{figure*}

\section{Results}

\subsection{Architecture Search}

The architecture search stage is the most computationally expensive.  We are hence interested in both the search time and performance, both of which are illustrated in Table~\ref{tab:time} for experiments 
with DARTS and PC-DARTS for models with 4 layers and 16 channels \((L=4,C=16)\). In DARTS case, the batch size is set to the largest possible given GPU memory constraints. The use of partial connections improves on search time by approximately 50\% while regularisation results in improved accuracy.  
Performance also translates well from the training partition to the development partition. Illustrations of the resulting normal and reduction cell operations are shown in Figure~\ref{fig:cellIllustrations}.

\begin{table}[t]
      \caption{A comparison of DARTS and PC-DARTS models with \(L=4\) layers and \(C=16\) channels.  Results in terms of processing efficiency (GPU-days) and accuracy for ASVspoof 2019 LA training and development partitions.   
      }
      \label{tab:time}
      \centering
      \resizebox{\columnwidth}{!}{%
      \setlength\tabcolsep{2pt}
      \begin{tabular}{ccccc}
        \toprule
               &  &  Search Cost    & \multicolumn{2}{c}{Best Architecture}  \\
        \textbf{Model size} &\textbf{Systems}&\textbf{GPU-days}&\textbf{Train Acc}&\textbf{Dev Acc}\\
        \midrule

         $(L=4,$ & DARTS&   0.29   &  98.80  &   97.21  \\
           $C=16)$&PC-DARTS &   0.15   &   99.97 &   100   \\
        \bottomrule
      \end{tabular}
      }
    \end{table}
    
\subsection{Train from Scratch}

Table~\ref{tab:model_size} shows results for a set of different PC-DARTS configurations (column 1) and number of parameters (column 2).
min-tDCF and EER results are shown for both the development partition (columns~3 and~4) and evaluation partition (columns~5 and~6). According to the primary min-tDCF metric, the best performing model has 16 layers and 64 initial channels. For the evaluation partition, it delivers a min-tDCF of 0.0914 and an EER of 4.96\%.
The second best model with 4 layers and 16 initial channels delivers a min-tDCF of 0.0992 and an EER of 5.53\%. This is achieved with 7.37M fewer parameters. Performance for the smallest model is substantially degraded in terms of min-tDCF, albeit if the EER is still respectable. The largest tested model size offers no benefit in terms of performance which is likely the result of over-fitting to training data.

 \begin{table}[!t]
 \vspace{-0.3cm}
 \caption{Number of parameters and results for a selection of different PC-DARTS models.  Results for the ASVspoof 2019 LA database. }
 \label{tab:model_size}
 \centering
 \resizebox{\columnwidth}{!}{
 \setlength\tabcolsep{2pt}
\begin{tabular}{lccccc}
\toprule
\multicolumn{1}{l}{}   & \multicolumn{1}{c}{}  & \multicolumn{2}{c}{Dev}  & \multicolumn{2}{c}{Eval}\\
\multicolumn{1}{c}{\textbf{Model size}}  &  \multicolumn{1}{c}{\textbf{Params}} &
\multicolumn{1}{c}{\textbf{min-tDCF}} & \multicolumn{1}{c}{\textbf{EER}} & \multicolumn{1}{c}{\textbf{min-tDCF}} & \multicolumn{1}{c}{\textbf{EER}} \\ \hline
\multicolumn{1}{l}{${\left(L=2, C=4\right)}$}& \multicolumn{1}{c}{0.007M}   &   
\multicolumn{1}{c}{0.0004}  &  \multicolumn{1}{c}{0.04}  &  \multicolumn{1}{c}{0.1244}  & \multicolumn{1}{c}{5.80}  \\
\multicolumn{1}{l}{${\left(L=4, C=16\right)}$}& \multicolumn{1}{c}{0.14M}   & 
\multicolumn{1}{c}{0}  &  \multicolumn{1}{c}{0}  &  
\multicolumn{1}{c}{0.0992}  & \multicolumn{1}{c}{5.53}                  \\
\multicolumn{1}{l}{${\left(L=8, C=32\right)}$} &  \multicolumn{1}{c}{0.97M}  &  
\multicolumn{1}{c}{0.00004} & \multicolumn{1}{c}{0.002}              & 
\multicolumn{1}{c}{0.1177} &  \multicolumn{1}{c}{4.87}                  \\
\multicolumn{1}{l}{${\left(L=16, C=64\right)}$} &  \multicolumn{1}{c}{7.51M} & \multicolumn{1}{c}{0}  & \multicolumn{1}{c}{0}  &    
\multicolumn{1}{c}{0.0914} &  \multicolumn{1}{c}{4.96} \\
\multicolumn{1}{l}{${\left(L=24, C=64\right)}$} &  \multicolumn{1}{c}{10.57M} & 
\multicolumn{1}{c}{0.0001}  & \multicolumn{1}{c}{0.039}  &   
\multicolumn{1}{c}{0.1045} &  \multicolumn{1}{c}{5.51} \\\bottomrule

\end{tabular}
}
\end{table}

\subsection{Comparison to competing systems}

Table~\ref{tab:baselines} shows 
a comparison of results to top-performing systems reported in the literature and the two ASVspoof baselines~\cite{todisco2019asvspoofbaselines}. The best (16,64) model achieves substantially better performance than the two ASVspoof baselines and also outperforms all but two others, both Res2Net models.  Even then, the differences in terms of min-tDCF are modest (even if greater in terms of EER). Our second best model, with 85\% fewer parameters than the best Res2Net model,  remains competitive. These are satisfying results and are the first to show that anti-spoofing models whose \emph{architecture and parameters} are learned automatically can compete with models designed with models designed with far greater human effort.

    \begin{table}[t]
      \caption{A performance comparison between PC-DARTS models and competing state-of-the-art systems reported in the literature.  Results for the ASVspoof LA evaluation partition.}
      \label{tab:baselines}
      \centering
      \resizebox{\columnwidth}{!}{%
      \setlength\tabcolsep{3pt}
      \begin{tabular}{c c c c c c}
        \toprule
        {\textbf{Systems}}&{\textbf{Features}} &{\textbf{min-tDCF}}&{\textbf{EER}}&{\textbf{Params}}\\
        \midrule
         Res2Net\cite{li2020replay}&CQT &0.0743 & 2.50& 0.96M \\
         Res2Net\cite{li2020replay}&LFCC &0.0786 & 2.87& 0.96M \\
         PC-DARTS $(16,64)$&LFCC &0.0914 & 4.96& 7.51M \\
         PC-DARTS $(4,16)$&LFCC &0.0992 & 5.53& 0.14M \\
         LCNN\cite{lavrentyeva2019stc}\cite{liu2019adversarial} &LFCC   &   0.1000   &   5.06&   10M\\ 
         LCNN\cite{lavrentyeva2019stc}\cite{liu2019adversarial} &LPS   &   0.1028   &   4.53&   10M\\ 
         LFCC-GMM\cite{todisco2019asvspoofbaselines}   &LFCC&   0.2116   &   8.09&   - \\  
         Res2Net\cite{li2020replay} &LPS&0.2237 & 8.78& 0.96M \\
         CQCC-GMM\cite{todisco2019asvspoofbaselines}   & CQCC &   0.2366   &   9.57&   - \\
         Deep Res-Net\cite{alzantot2019deep} &LPS  &   0.2741&   9.68&   0.31M  \\
        \bottomrule
      \end{tabular}
      }
    \end{table}

\section{Conclusions}
This paper reports what is, to the best of our knowledge, the first successful application of neural architecture search (NAS) to the spoofing detection problem.  We show that partially connected differentiable architecture search (PC-DARTS) is able to learn complex neural architectures from a fixed set of candidate operations.
Architectures learned with PC-DARTS can be optimised using backpropagation and with hardware acceleration, meaning that even complex convolutional and residual networks can be learned automatically.  

The performance of the resulting models is competitive with the state of the art. Our best performing model achieves a min-tDCF of 0.09 
for the ASVspoof 2019 Logical Access database, a result outdone only by a Res2Net system, and even then only by a modest margin.  Given that our result was generated by a network whose architecture and parameters are all learned automatically, instead of from many hours of manual optimisation, this is a satisfying result. Our second-best system which achieves a min-tDCF of 0.1 has 85\% fewer parameters than the best performing Res2Net system. With these results, we are encouraged to pursue PC-DARTS further.  The obvious next step is to apply PC-DARTS directly to raw signal inputs.  Other directions include the use of PC-DARTS as a full end-to-end solution to both spoofing detection and automatic speaker verification.

\section{Acknowledgements}

This work is supported by the TReSPAsS-ETN project funded from the European Union’s Horizon 2020 research and innovation programme under the Marie Skłodowska-Curie grant agreement No.860813. It is also supported by the ExTENSoR project funded by the
French Agence Nationale de la Recherche (ANR).

\balance
\bibliographystyle{IEEEtran}
\bibliography{mybib}

\end{document}